\documentclass[letterpaper, 10 pt, conference]{ieeeconf} 
\IEEEoverridecommandlockouts            
\usepackage{epsfig} 
\usepackage{mathptmx} 
\usepackage{times} 
\usepackage{amsmath} 
\usepackage{amssymb}  
\usepackage{cite}
\usepackage{graphicx} 
\usepackage[hidelinks]{hyperref}
\usepackage{multirow,color}

\title{\LARGE \bf
Improving Disturbance Estimation and Suppression via Learning among Systems with Mismatched Dynamics
}

\author{Harsh Modi$^{1}$, Zhu Chen$^{2}$, Xiao Liang$^{3,*}$, and Minghui Zheng$^{1,*}$
\thanks{\textcopyright 2024 IEEE.  Personal use of this material is permitted.  Permission from IEEE must be obtained for all other uses, in any current or future media, including reprinting/republishing this material for advertising or promotional purposes, creating new collective works, for resale or redistribution to servers or lists, or reuse of any copyrighted component of this work in other works.}
\thanks{$^{1}$Harsh Modi ({\tt\small harsh.modi@tamu.edu}, current Ph.D. student) and Minghui Zheng ({\tt\small mhzheng@tamu.edu}, faculty member) are with the Department of Mechanical Engineering, Texas A\&M University, TX 77843.}%
\thanks{Zhu Chen ({\tt\small zhuchen@buffalo.edu}, former Ph.D. student) was with Mechanical and Aerospace Engineering, University at Buffalo, NY 14228}
\thanks{$^{3}$Xiao Liang ({\tt\small xliang@tamu.edu}, faculty member) is with the Department of Civil \& Environmental Engineering, Texas A\&M University, TX 77843. }%
\thanks{This work is supported by U.S. National Science Foundation (Grants: No. 2030375/2422579 and No. 2046481/2422698). Correspondence to Minghui Zheng and Xiao Liang.}
}

\begin{document}
\maketitle
\thispagestyle{plain}
\pagestyle{plain}
\begin{abstract}
Iterative learning control (ILC) is a method for reducing system tracking or estimation errors over multiple iterations by using information from past iterations. The disturbance observer (DOB) is used to estimate and mitigate disturbances within the system, while the system is being affected by them. ILC enhances system performance by introducing a feedforward signal in each iteration. However, its effectiveness may diminish if the conditions change during the iterations. On the other hand, although DOB effectively mitigates the effects of new disturbances, it cannot entirely eliminate them as it operates reactively. Therefore, neither ILC nor DOB alone can ensure sufficient robustness in challenging scenarios. This study focuses on the simultaneous utilization of ILC and DOB to enhance system robustness. The proposed methodology specifically targets dynamically different linearized systems performing repetitive tasks. The systems share similar forms but differ in dynamics (e.g. sizes, masses, and controllers). Consequently, the design of learning filters must account for these differences in dynamics. To validate the approach, the study establishes a theoretical framework for designing learning filters in conjunction with DOB. The validity of the framework is then confirmed through numerical studies and experimental tests conducted on unmanned aerial vehicles (UAVs). Although UAVs are nonlinear systems, the study employs a linearized controller as they operate in proximity to the hover condition. A video introduction of this paper is available via this \href{https://zh.engr.tamu.edu/wp-content/uploads/sites/310/2024/02/ILCDOB_v3f.mp4}{\textbf{\textcolor{blue}{link}}}.
\end{abstract}
\section{Introduction}

Deploying safety-critical robotic systems such as unmanned aerial vehicles (UAVs) in the vicinity of human presence requires them to possess robustness against external factors, such as wind disturbances. These disturbances can significantly impact the trajectory of the UAVs, posing potential dangers to external subjects \cite{wind_effects} \cite{UAV_dangerous}. Therefore, it is crucial for these robotic systems to estimate and mitigate the effects of disturbances to ensure the necessary level of safety. Also, various tasks may require the use of dynamically different systems in disturbance-prone environments.

Iterative learning control (ILC) is effective in reducing system error over multiple iterations in repetitive tasks, thereby enhancing performance in each iteration. It has been successfully applied in various applications, including manipulator-based robotics systems \cite{ILC_survey}. Recently, ILC has been utilized to improve UAV's trajectory tracking performance \cite{ILC_UAV} using optimization-based filter designs. Researchers in \cite{ILC_parameter_determination} implemented parameter determination-based ILC for robotic manipulators, while \cite{ILC_sliding_mode} combined ILC with sliding mode control to enhance the trajectory tracking for UAVs. 

The disturbance observer (DOB) has been widely used to enhance the robustness of the controller against external disturbances. The article \cite{DOB_survey} provides an overview of advancements in DOB from 1985 to 2020. DOB-based controllers have been employed to compensate for unknown disturbances in small UAV systems \cite{DOB_UAV}. In \cite{DOB_acc}, the authors used DOB with a disturbance rejection signal in the form of acceleration, which is similar to the force exerted on the UAV due to disturbances like wind. \cite{DOB_FTDO} used finite-time disturbance observer to mitigate disturbance effects for quadrotor UAVs, and \cite{DOB_LDOB} developed a linear dual disturbance observer to improve UAV trajectory tracking. DOB has also been used to enhance the robustness of fixed-wing UAVs \cite{DOB_fixed_wing}, and in \cite{DOB_nonlinear}, a disturbance observer was designed for nonlinear and nonautonomous systems.

The objective of this research is to combine the advantages of both the ILC and DOB to improve system robustness. Specifically, this study focuses on increasing the robustness of UAV trajectory tracking against external disturbances, while simultaneously estimating the disturbance present in the environment. When UAVs follow the same trajectories within a relatively short period, it can be assumed that the disturbances will not vary significantly. In such cases, the benefits of ILC can be leveraged to proactively compensate for repetitive errors in trajectory tracking caused by disturbance or controller limitations. However, ILC alone cannot account for changing conditions, which can be addressed by incorporating DOB. By combining ILC and DOB, we can utilize the proactivity of ILC and the ability of DOB to adapt to new disturbances. 

Many studies have explored this direction. \cite{ILC_DOB_harness} employed ILC along with DOB to account for non-repetitiveness in the disturbances. In \cite{ILC_DOB_wafer}, the performance of ILC was enhanced with DOB for wafer scanning systems. \cite{ILC_DOB_excavation} utilized the combined ILC and DOB to reject near-repetitive disturbances in excavation operations. \cite{ILC_DOB_machine_tools} combined ILC with DOB to improve the robustness of machine tool feed drives. \cite{ILC_DOB_mismatched} improved the closed-loop performance using ILC based on DOB. \cite{ILC_DOB_rehabilitation} employed ILC with DOB for rehabilitation. In \cite{ILC_DOB_nonlinear}, ILC was combined with disturbance estimation for unmatched model uncertainties and matched disturbances. However, all these studies utilized the same system in each iteration, limiting the robustness to a single system. 

Some research has focused on implementing iterative learning in dynamically different systems. \cite{ILC_hetero_initial_condition} extended the capability of ILC in heterogeneous systems with different initial conditions. \cite{ILC_hinf} used a transfer learning approach to transfer input learning to a different system, but it was specifically targeted at improving trajectory tracking and did not include disturbance estimation. This study is an extension of \cite{zhu_ILC_hetero}, where we combine ILC with DOB for dynamically different systems. In \cite{zhu_ILC_hetero}, the authors designed learning filters for dynamically different systems, allowing each subsequent system to learn from the errors and learning signal of the previous system. The design was based on guaranteeing that the learning-based trajectory tracking error of each system would be smaller than the system without learning. However, explicit disturbance rejection was not included, which is a focus of the present study, along with the learning.

The main contributions of this study are as follows: we explicitly incorporate DOB as part of the ILC update process and consider differences in system dynamics among different systems to enable learning. To the best of our knowledge, this is the first attempt to implement learning with DOB for systems with mismatched dynamics. Furthermore, the designed methodology has undergone rigorous verification and validation through simulations and experiments. In the current study, the disturbance rejection and learning framework is implemented as follows: (1) All systems operate with an underlying PID baseline controller that remains unmodified in this study. As the systems operate near hover conditions, the linearization approximation holds. (2) All systems utilize a DOB algorithm in conjunction with the PID controller to estimate and reject the disturbances. However, DOB alone cannot fully compensate for disturbance effects. Therefore, information regarding the tracking error is passed to the next system. (3) The next system utilizes this information to generate a learning signal, which aids in improved disturbance estimation and rejection compared to non-learning scenarios. 

\begin{figure}[htbp]
\centering 
{\includegraphics[width=0.36\textwidth]{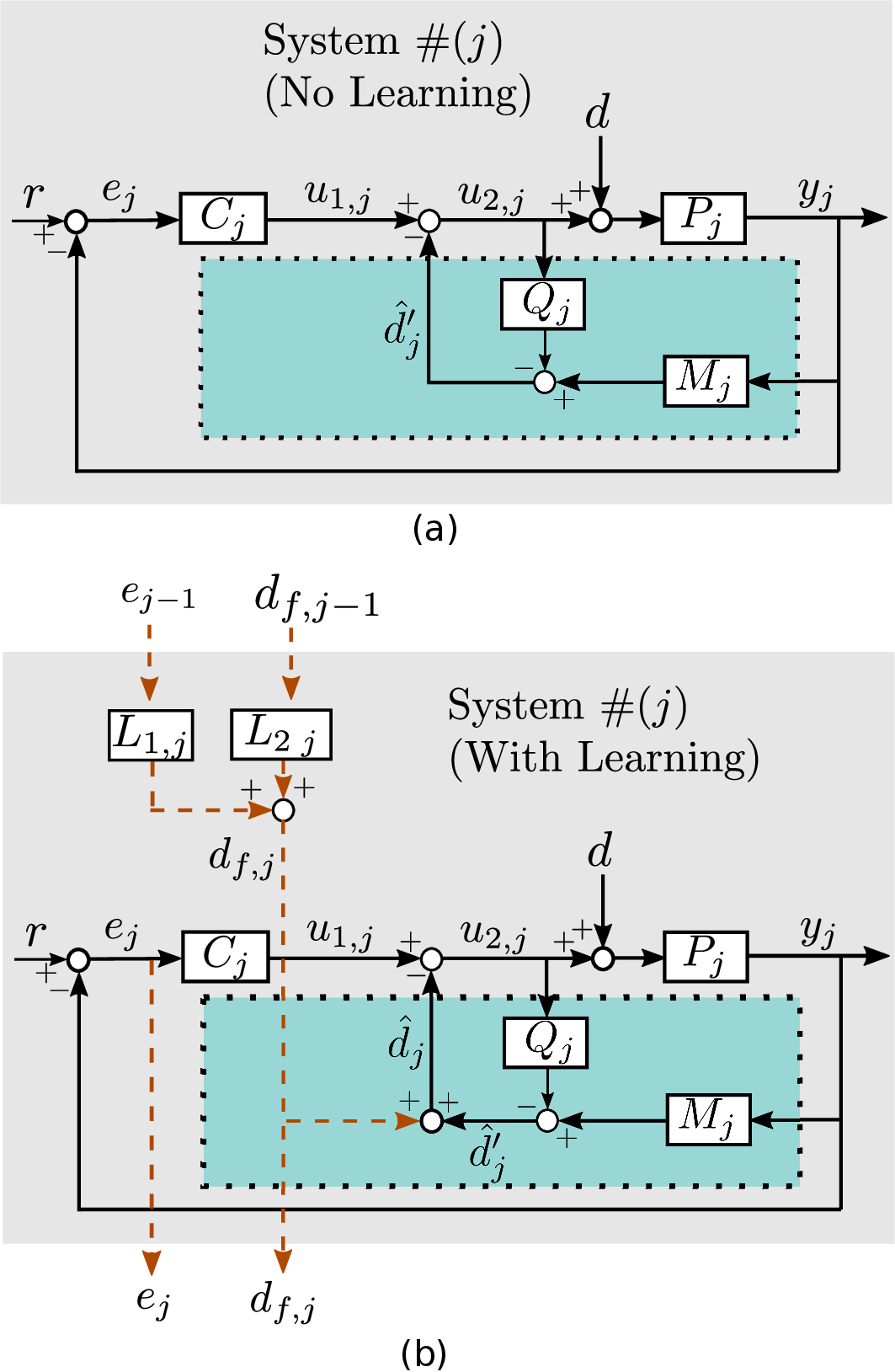}}
\caption{System block diagram with (a) a basic DOB framework (b) an iterative learning framework along with DOB}
\label{fig:basic_dob_ilc_framework_combined}
\vspace{-25pt}
\end{figure}

The rest of the paper is organized as follows: Section \ref{section:framework} establishes the theoretical framework and describes the design of the learning filters.
Section \ref{section:validation} presents the simulation and the experimental results. Section \ref{section:conclusion} concludes the article. Please note that throughout the paper, the term ``UAV" indicates the particular hardware used in the experiment while the term ``system" is used to describe the general order of the learning iterations.

\section{Learning Framework}
\label{section:framework}
\subsection{Variable Definition and Standard DOB Basics}
We first introduce notations that will be used in our framework. We denote the signals as follows: $r(k)$ as the reference input, $y(k)$ as the output, $e(k)$ as the tracking error, $d(k)$ and $\hat{d}(k)$ as the disturbance and its estimate, $u_1(k)$ as the control signal directly generated by the baseline feedback control, $u_2(k)$ as the modified control signal that is sent to the plant. All these signals are time series. We also introduce the following notations for different subsystems: $P(z)$ as the plant, $C(z)$ as the baseline controller, $Q(z)$ as a low-pass filter, $M(z)$ as the plant inverse, $L_1(z)$ and $L_2(z)$ as learning filters; all of these are transfer functions in discrete time. In addition, we use $j$ to index different systems; and use the prime symbol ($'$) to distinguish signals in systems without learning. For ease of reading, we will omit $k$ in signals and $z$ in transfer functions.

We now introduce the standard DOB and explain how it works. As shown in Fig. \ref{fig:basic_dob_ilc_framework_combined}(a), the DOB is added to System $j$. It consists of a plant inverse $M$ and a low-pass filter $Q$, as highlighted by the dotted box. When a disturbance $d$ is present, the DOB can provide a disturbance estimate $\hat{d}$ which will be subtracted from $u_1$ to cancel. Ideally, if a plant inverse can be accurately obtained, and the intrinsic delay in $P$ is small, $\hat{d}$ would be close to $d$ so that the disturbance can be suppressed. However, it is difficult to accurately estimate the plant inverse and the delays exist in dynamic systems. These limitations of the DOB can be addressed by using a learning approach across multiple iterations, as proposed in this study. The subsequent section elaborates on the development of this learning framework over DOB for dynamically different systems. 

\subsection{Iterative Learning with DOB Framework}

\begin{figure}[htbp]
         \centering {\includegraphics[width=0.42\textwidth]{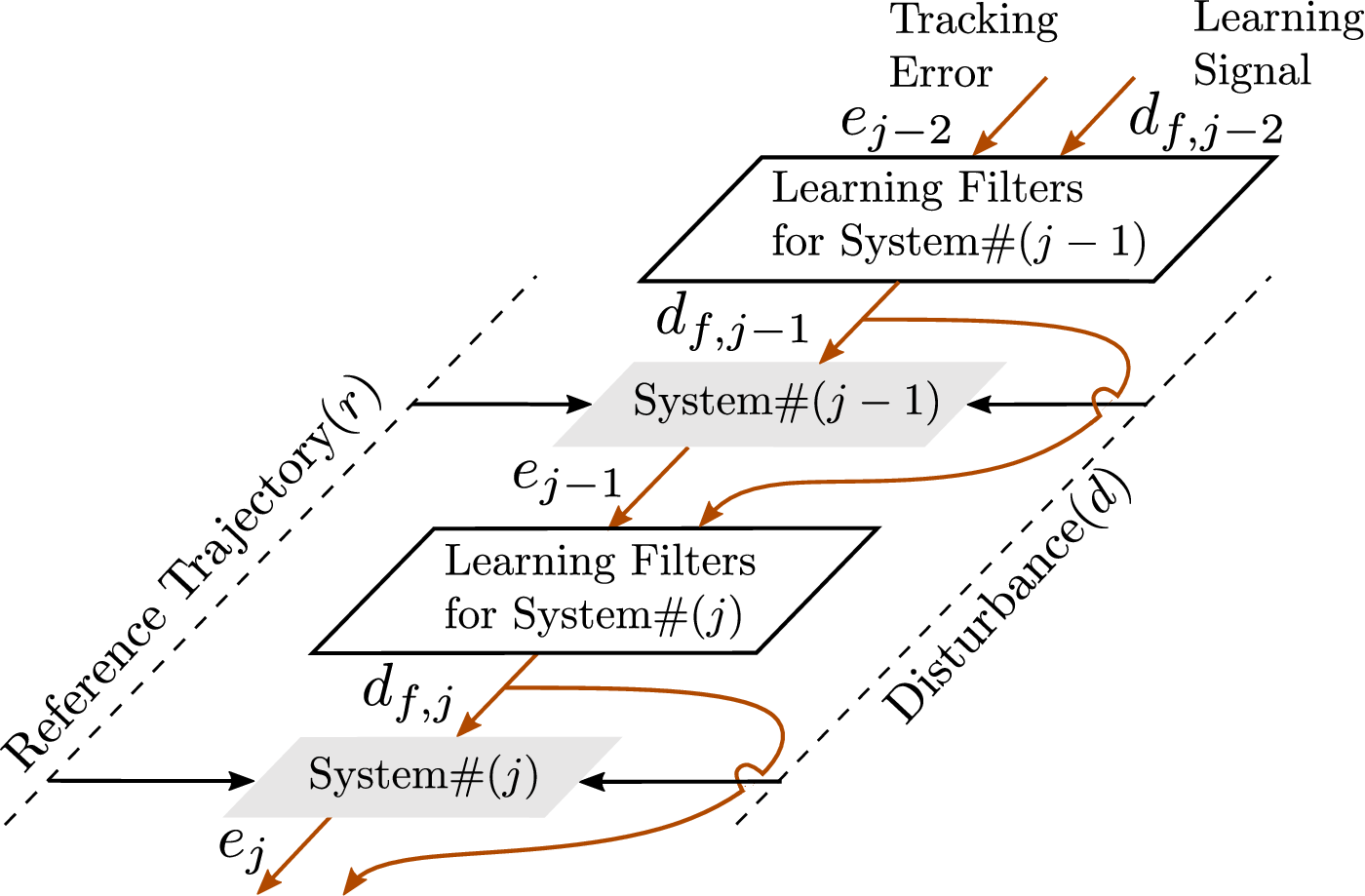}}
      \caption{Learning relationship among systems}
      \label{fig:learning_order}
      \vspace{-15pt}
\end{figure}

In this subsection, we will introduce the iterative learning framework with DOB. Fig. \ref{fig:basic_dob_ilc_framework_combined}(b) shows the detailed system block diagram with the learning framework introduced, whereas Fig.~\ref{fig:learning_order} shows the overall learning relationship between two systems having different dynamics, i.e. System $\#(j{-}1)$ and System $\#(j)$. The System $\#(j{-}1)$'s learning signal ($d_{f,j-1}$) and its trajectory tracking error ($e_{j-1}$) can qualify the accuracy of its disturbance estimation. Hence, $d_{f,j}$ is generated using $e_{j-1}$ and $d_{f,j-1}$ and to-be-deisgned learning filters for System $\#(j)$. We propose two different learning filters $L_{1,j}$ and $L_{2,j}$ respectively for $e_{j-1}$ and $d_{f,j-1}$ as illustrated in Fig. \ref{fig:basic_dob_ilc_framework_combined}(b). As the framework aims to improve the system's disturbance estimation and suppression capability as well as its trajectory-tracking performance, the learning signal is added to the disturbance estimate ($\hat{d}'_j$) from the DOB. It is important to note that learning aims to improve the performance of the system with learning compared to its performance without learning. We do not compare the performance of the system $\#(j)$ with system $\#(j{-}1)$ as each system's dynamics can affect individual performance. 

The derivation is based on the following assumptions: 1. We consider that all the systems are linear time-invariant (LTI) as a large number of non-linear time-invariant systems can be linearized near the equilibrium point. 2. As the research is aimed at reducing the effects of disturbances on repetitive tasks (such as industrial assembly or delivery robots), we consider they follow the same trajectories and are subject to similar disturbances 3. The current research aims at slowly time-varying or stationary reference signals.

In the following paragraphs, we will establish the relationship between the tracking error with learning ($e_j$) and the tracking error without learning ($e_j'$) after stating some system parameters' definitions.

\noindent
\textbf{System Parameters:}
Based on the system block diagram in Fig. \ref{fig:basic_dob_ilc_framework_combined}(b), $G_{r,j}$ (dynamics from reference signal $r$ to output $y_j$), $G_{d,j}$ (dynamics from disturbance $d$ to output $y_j$), and $G_{f,j}$ (dynamics from learning signal $d_{f,j}$ to output $y_j$) can be described by: 
 \begin{equation}
     G_{r,j} = [1-Q_j+P_j(M_j + C_j)]^{-1}P_jC_j
 \end{equation}  
 \begin{equation}
    \label{equation:Gdj}
     G_{d,j} = [1-Q_j+P_j(M_j + C_j)]^{-1}P_j(1-Q_j)
 \end{equation} 
 \begin{equation}
    \label{equation:Gfj}
    G_{f,j} = [1-Q_j+P_j(M_j + C_j)]^{-1}(-P_j)
\end{equation} 

\noindent
\textbf{Establishing Relationship between $e_j$ and $e'_j$:}
Using the system parameters defined above, the output of system $\#(j)$ with learning is given by:

\begin{equation}
    \label{equation:yj}
    y_j = G_{r,j}\{r\} + G_{d,j}\{d\} + G_{f,j}\{d_{f,j}\}
\end{equation}

\noindent and the output of the system $\#(j)$ without learning is given by: 
\begin{equation}
    \label{equation:yj_prime}
    y'_j = G_{r,j}\{r\} + G_{d,j}\{d\}
\end{equation}
where the notation \{\} indicates that the signal inside is sent to a system which can be represented by the outside transfer function. 

In order to establish the relationship between $e_j$ and $e'_{j}$, let us expand $e_j-e'_{j}$:
\begin{equation}
    e_j-e'_j=(r-y_j)-(r-y'_j)=y'_j-y_j
\end{equation}

\noindent Now, using Eq. (\ref{equation:yj}) and Eq. (\ref{equation:yj_prime}):
\begin{equation}
    \label{equation:error_compare_first}
        e_j=e'_j+y'_j-y_j=e'_j-G_{f,j}\{d_{f,j}\}
\end{equation}

As discussed earlier, the learning signal of the current system ($d_{f,j}$) is based on $e_{j-1}$, and $d_{f,j-1}$ and the respective to-be-designed learning filters $L_{1,j}$, and $L_{2,j}$. Hence:
\begin{equation}
        \label{equation:learning_signal}
        d_{f,j} = L_{1,j}\{e_{j-1}\}+L_{2,j}\{d_{f,j-1}\}
\end{equation}

\noindent Using this, we expand Eq. (\ref{equation:error_compare_first}) further as:
\begin{equation}
    \label{equation:error_compare_second}
    \begin{split}
        e_j=&e'_j-G_{f,j}(L_{1,j}\{e_{j-1}\}+L_{2,j}\{d_{f,j-1}\})
    \end{split}
\end{equation}

\noindent Now, $e_{j-1}$ can be expressed using Eq. (\ref{equation:yj}) as:
\begin{equation}
    \label{equation:tracking_error_j-1_learning}
    \begin{split}
    e_{j-1}&=r-y_{j-1}\\&=(1-G_{r,j-1})\{r\}
    -G_{d,j-1}\{d\}-G_{f,j-1}\{d_{f,j-1}\}
    \end{split}
\end{equation}

\noindent Using this in Eq. (\ref{equation:error_compare_second}) and after some simplification:

\begin{equation}
    \label{equation:intermediate_1}
    \begin{split}
        e_j=&e'_j-G_{f,j}L_{1,j}(1-G_{r,j-1})\{r\}+G_{f,j}L_{1,j}G_{d,j-1}\{d\}\\&+G_{f.j}(L_{1,j}G_{f,j-1}-L_{2,j})\{d_{f,j-1}\}\\
    \end{split}
\end{equation}

Eq. (\ref{equation:intermediate_1}) contains variables $r$, $d$, and $d_{f,j-1}$ apart from $e'_j$. In order to effectively establish a relationship between $e_j$ and $e'_j$, let us try to reduce the number of variables in the equation. Using Eq. (\ref{equation:yj_prime}), we can express $e'_j$ as:

\begin{equation}
\label{equation:tracking_error_j-1_dob}
    e'_j=r-y'_j=(1-G_{r,j})\{r\}-G_{d,j}\{d\}
\end{equation}

\noindent Hence, $d$ can be expressed in terms of $r$ and $e'_j$ as:
\begin{equation}
    \label{equation:disturbance}
    d=G^{-1}_{d,j}(r-G_{r,j}\{r\}-e'_j)
\end{equation}
\noindent substituting this in Eq. (\ref{equation:intermediate_1}) and with some re-arrangements, we get
\begin{equation}
    \label{equation:error_j_with_reference}
    \begin{split}
        e_j=&(1-G_{f,j}L_{1,j}G_{d,j-1}G^{-1}_{d,j})\{e'_j\}\\&-G_{f,j}L_{1,j}((1-G_{r,j-1})-G_{d,j-1}G^{-1}_{d,j}(1-G_{r,j}))\{r\}\\&+G_{f,j}(L_{1,j}G_{f,j-1}-L_{2,j})\{d_{f,j-1}\}
    \end{split}
\end{equation} 
Considering that the trajectory tracking controller $C$ is well designed such that
\begin{small}
\begin{equation}
\label{equation:Gr_unity}
G_r(j\omega)=\frac{C(j\omega)P(j\omega)}{1+C(j\omega)P(j\omega)}\approx1
\end{equation}
\end{small}
for $\omega<\omega_0$,
\noindent where $\omega_0$ is the desired unity-gain bandwidth of the closed-loop system $G_r$.

As seen in a later section, for the UAVs used in this study, the gain of the transfer function $G_r$ is close to 0 dB and the phase is close to $0^\circ$ for $\omega<1\ rad/s$. Considering Eq. (\ref{equation:Gr_unity}) and with the presence of $(1{-}G_{r,j-1})$ and $(1{-}G_{r,j})$, the following term 
\begin{equation}
\label{equation:Grj_zero}
-G_{f,j}L_{1,j}((1-G_{r,j-1})-G_{d,j-1}G^{-1}_{d,j}(1-G_{r,j}))\{r\} \approx 0
\end{equation}
for stationary or slowly time-varying trajectories.

With Eq. (\ref{equation:Grj_zero}), we can re-write Eq. (\ref{equation:error_j_with_reference}) as:
\begin{equation}
    \label{equation:error_j_without_reference}
    \begin{split}
    e_j=&(1-G_{f,j}L_{1,j}G_{d,j-1}G^{-1}_{d,j})\{e'_j\}+\\&G_{f,j}(L_{1,j}G_{f,j-1}-L_{2,j})\{d_{f,j-1}\}
    \end{split}
\end{equation}
This relationship between $e_j$ and $e_j'$ will be used in designing the learning filters.

\subsection{Learning Filter Design}
In this subsection, we will introduce the design of the learning filters, i.e., $L_{1,j}$ and $L_{2,j}$, such that $e_j$ can be reduced compared to $e_j'$. For simplification, we introduce the following notations:
\begin{equation}
\label{equation:te1j_te2j_def}
\begin{split}
    T_{e_{1,j}}=&(1-G_{f,j}L_{1,j}G_{d,j-1}G^{-1}_{d,j})\\
    T_{e_{2,j}}=&G_{f,j}(L_{1,j}G_{f,j-1}-L_{2,j})
\end{split}   
\end{equation}
Using this, Eq. (\ref{equation:error_j_without_reference}) can be re-written as: 
\begin{equation}
    \label{equation:ej_tej}
    e_j=T_{e_{1,j}}\{e'_j\}+T_{e_{2,j}}\{d_{f,j-1}\}
\end{equation}

To explicitly analyze robustness, we denote the modeling uncertainty as $\Delta_j$ and the relationship between the actual plant ($P_j$) and the identified plant model ($\hat{P}_j$) as: 
\begin{equation}
    \label{equation:inaccuracy_model}
    P_j=(1+\Delta_j) \hat{P}_j
\end{equation}

In the following paragraphs, we will introduce the learning filter design in a theorem-proof format. We also split our proof into two cases: without and with modeling uncertainties. We use $\| \cdot \|$ to denote the 2-norm of a signal and the H-infinity norm of a system. That is, 
\begin{equation}
\|s\|=(\sum_0^\infty [s(k)]^2 )^{1/2}~\text{and}~     \|G\|=\max_\omega|G(j\omega)|
\end{equation}
where $s$ is a discrete-time signal and $G$ is a LTI transfer function. 
\\
\noindent \textbf{Theorem:}
The following learning filters \begin{equation}
\label{equation:L1j}
    L_{1,j}=(\hat{G}_{f,j}\hat{G}_{d,j-1})^{-1}\hat{G}_{d,j}
\end{equation} and 
\begin{equation}
\label{equation:L2j}
    L_{2,j}=(\hat{G}_{f,j}\hat{G}_{d,j-1})^{-1}\hat{G}_{d,j}\hat{G}_{f,j-1}
\end{equation} can ensure $||e_{j}||<||e'_{j}||$ if \begin{equation}
||\Delta_{j-1}||<||(1+P_{j-1}C_{j-1})/2||
\end{equation}

\noindent 
\textbf{Proof:}
With the learning filters in Eq. (\ref{equation:L1j}) and Eq. (\ref{equation:L2j}), Eq. (\ref{equation:te1j_te2j_def}) can be expanded as:
\begin{equation}
\label{equation:te1j_te2j_with_learning_filters}
\begin{split}
    T_{e_{1,j}}=&(1-G_{f,j}[\hat{G}_{f,j}\hat{G}_{d,j-1})^{-1}\hat{G}_{d,j}]G_{d,j-1}G^{-1}_{d,j})\\
    T_{e_{2,j}}=&G_{f,j}([\hat{G}_{f,j}\hat{G}_{d,j-1})^{-1}\hat{G}_{d,j}]G_{f,j-1}\\&-[\hat{G}_{f,j}\hat{G}_{d,j-1})^{-1}\hat{G}_{d,j}\hat{G}_{f,j-1}])
\end{split}   
\end{equation}

Now, let us analyze Eq. (\ref{equation:te1j_te2j_with_learning_filters}) in 2 different cases.\\

\noindent\textit{Case I: No modeling uncertainty}

When there is no modeling uncertainty, i.e., $\Delta_{j}=0$ and $\Delta_{j-1}=0$, we have $G_{f,j}=\hat{G}_{f,j}$, $G_{d,j}=\hat{G}_{d,j}$, $G_{f,j-1}=\hat{G}_{f,j-1}$, and $G_{d,j-1}=\hat{G}_{d,j-1}$. Hence, Eq. (\ref{equation:te1j_te2j_with_learning_filters}) reduces to
\begin{equation}
    \begin{split}
        &T_{e_{1,j}}=0\\
        &T_{e_{2,j}}=0
    \end{split}
\end{equation}

\noindent Therefore, with the derived learning filters, $e_j$ will converge to zero as per Eq. (\ref{equation:ej_tej}) when there is no modeling uncertainty.\\
\noindent \textit{Case II: With some modeling uncertainty:}

When there is some modelling uncertainty present, i.e. $||\Delta_{j}||\neq0$ or $||\Delta_{j-1}||\neq0$, we can simplify Eq. (\ref{equation:te1j_te2j_with_learning_filters})  using Eq. (\ref{equation:Gdj}), Eq. (\ref{equation:Gfj}), and Eq. (\ref{equation:inaccuracy_model}) as:
\begin{small}
\begin{equation}
\label{equation:Te1_errored}
    T_{e_{1,j}}=-\frac{\Delta_{j-1}}{1+P_{j-1} C_{j-1}}\\
\end{equation}
\end{small}
\begin{equation}
   \label{equation:Te2_errored}
    T_{e_{2,j}}=-\frac{\Delta_{j-1}}{1+P_{j-1} C_{j-1}}  G_{d,j-1}^{-1}G_{d,j}G_{f,j-1}
\end{equation}

\noindent Using Eq. (\ref{equation:tracking_error_j-1_dob}), Eq. (\ref{equation:ej_tej}), Eq. (\ref{equation:Te1_errored}), Eq. (\ref{equation:Te2_errored}), and with the approximation $||\Delta_j\Delta_{j-1}||\approx0$ for $\Delta_j$ and $\Delta_{j-1}$ of small gains, the error ``with learning" for System $\#(j)$ can be approximated as:
\begin{small}
\begin{equation}
\label{equation:ej_with_error}
    e_j\approx \frac{2\cdot\Delta_{j-1}Gd_{j}}{1+C_{j-1}P_{j-1}}\{d\}\approx \frac{-2\cdot\Delta_{j-1}}{1+C_{j-1}P_{j-1}}\{e'_j\}~(\text{for}~j\ge 2)
\end{equation}
\end{small}
\noindent As per Eq. (\ref{equation:ej_with_error}), to achieve $||e_j||<||e'_j||$, we need 
\begin{small}
\begin{equation}
\label{equation:delta_j-1_less_than_one}
\left\Vert\frac{-2\cdot\Delta_{j-1}}{1+C_{j-1}P_{j-1}}\right\Vert<1 
\end{equation}
\end{small}
\noindent As $1/(1+C_{j-1}P_{j-1})$ is the transfer function from $r$ to $e_{j-1}$, it will have a bounded gain for a stable closed-loop system. Hence, as per Eq. (\ref{equation:delta_j-1_less_than_one}), for $||\Delta_{j-1}||<||(1+P_{j-1}C_{j-1})/2||$, the $||e_j||$ is guaranteed to be less than $||e'_j||$. This result also confirms that if the system parameters are accurate (i.e. $\Delta_j=0$ for $\forall j$), then the error with learning will be 0 as proven earlier in case I. This is the end of the theorem proof. 

It is noted that the learning filters are explicitly dependent on model inverse; when the systems are not minimum-phase, we need to find their stable inverse approximation in the learning filter design \cite{DOB_non_minimum}.

\begin{table}[b]
    \vspace{-10pt}
    \caption{Specifications of the systems used}\vspace{-10pt}
    \label{table:UAVs}
    \begin{center}
    \begin{tabular}{|c|c|c|c|}
        \hline
          & \textbf{UAV}$\#(\mathbf{1})$ & \textbf{UAV}$\#(\mathbf{2})$ & \textbf{UAV}$\#(\mathbf{3})$\\
        \hline
        \textbf{Frame Brand} & F450 & S500 & Tarot 650\\
        \hline
        \textbf{Mass} & 0.921 kg & 1.001 kg & 1.234 kg\\
        \hline
        \textbf{Motor Distance from} & \multirow{2}{*}{228 mm }& \multirow{2}{*}{240 mm} & \multirow{2}{*}{318 mm}\\
        \textbf{Center of Mass} & & &\\
        \hline
        \textbf{Proportional Gain} & \multirow{2}{*}{3.0} & \multirow{2}{*}{1.2} & \multirow{2}{*}{1.9}\\
        \textbf{of PID controller} & & & \\
        \hline
    \end{tabular}
    \end{center}
\end{table}

\section{Simulation and Experimental Validation}
\label{section:validation}
In order to assess the effectiveness of the learning framework, the simulations and experiments have been performed on three dynamically different quadrotor UAVs as described in Table \ref{table:UAVs}. The quadrotor UAVs are inherently non-linear systems but are operating near their hover conditions using the linearized controller and hence we can assume that the linear system approximation holds here. It is to be noted that the proportional gains of the outer-loop PID controllers are kept different for each UAV to further vary their dynamics from each other. 
\begin{figure}[!htbp]
    \centering 
    \includegraphics[width=0.35\textwidth]{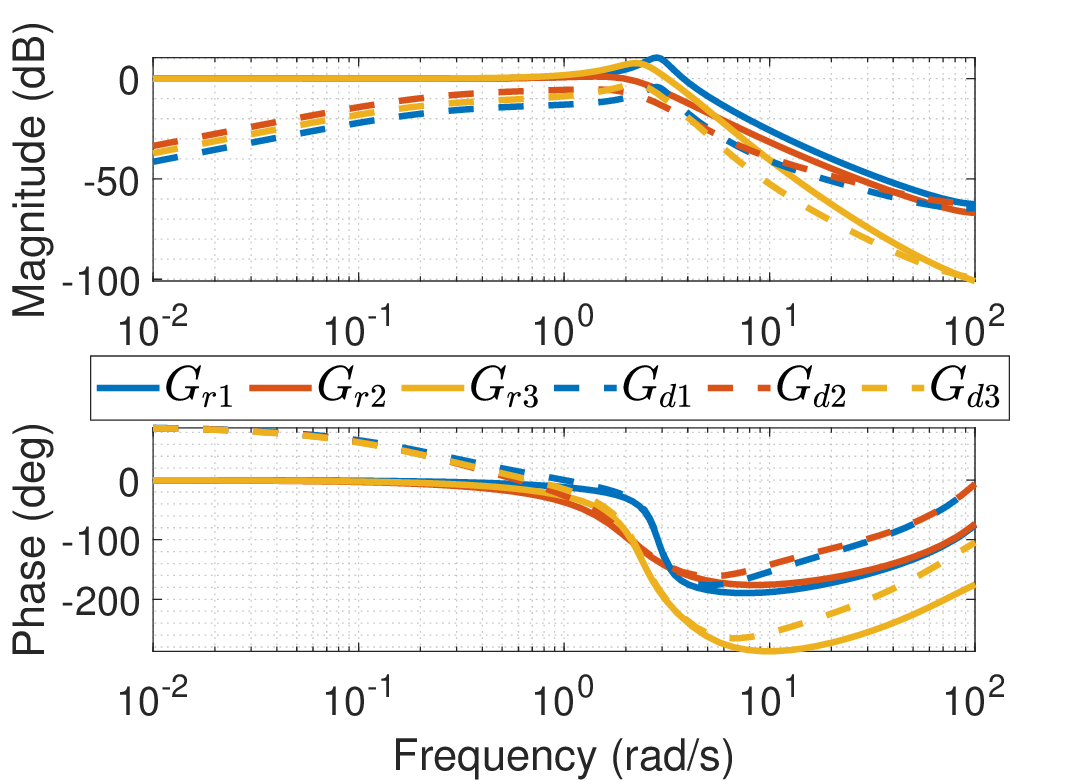}
    \caption{ Bode plots for $G_{r,j}$ and $G_{d,j}$}
    \label{fig:grd1_grd2_grd3}
    \vspace{-10pt}
\end{figure}  

\begin{figure}[!htbp]
    \centering \includegraphics[width=0.35\textwidth]{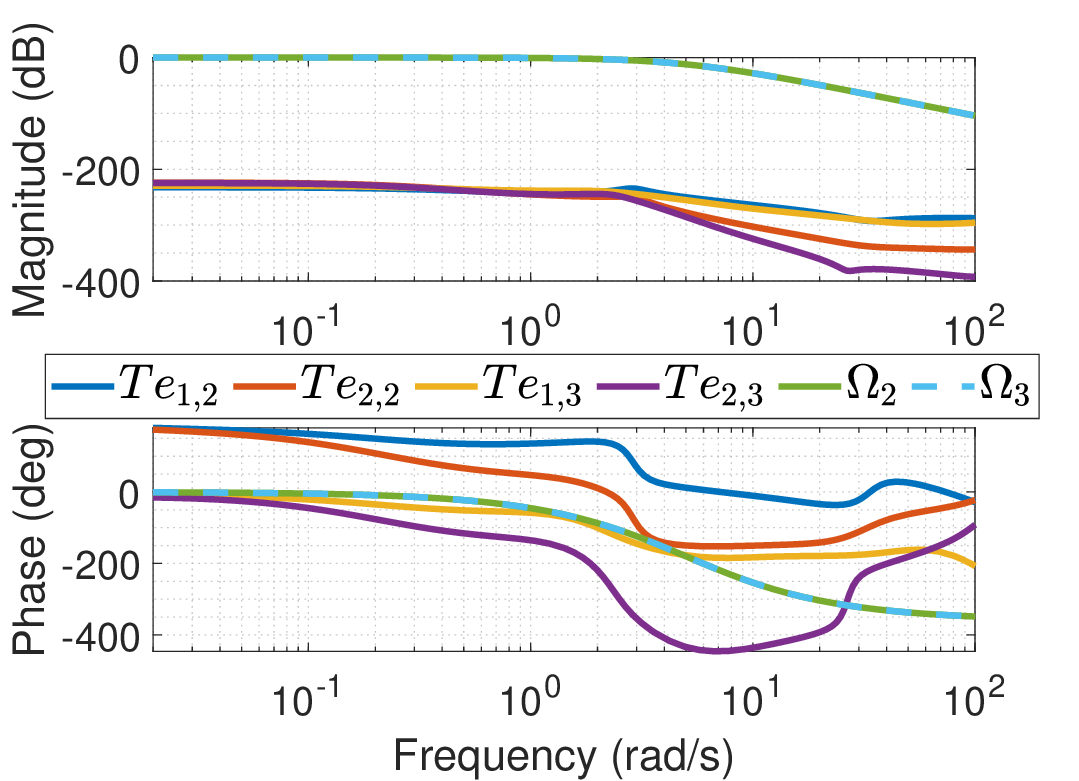}
    \caption{Bode plots for $T_{e_{1,j}}$, $T_{e_{2,j}}$, and $\Omega_j$}
    \label{fig:te_omega}
    \vspace{-5pt}
\end{figure}  

Using the identified transfer functions, we plot the Bode plots for the response of the systems given the reference ($G_{r,j}$) and the response of the systems given the disturbance ($G_{d,j}$) as shown in Fig. \ref{fig:grd1_grd2_grd3}. The $G_{d,j}$ for lower frequencies and very high frequencies have very low gain, indicating UAVs' inherent disturbance rejection capabilities in those frequencies. The middle frequencies (0.1 rad/s to 5 rad/s) are where the disturbances can have the maximum effects and hence we analyze the performance of the learning framework for the disturbances within this range.

\begin{figure}[!htbp]
    \centering \includegraphics[width=0.35\textwidth]{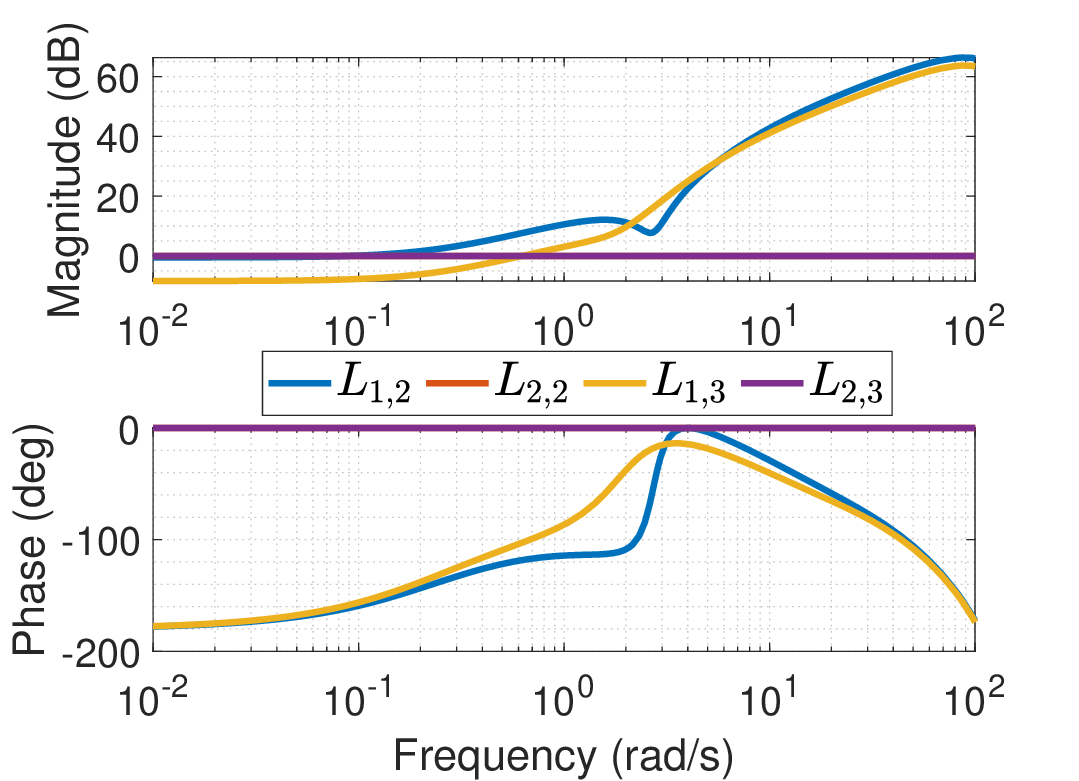}
    \caption{ Bode plots for $L_{1,j}$ and $L_{2,j}$}
    \label{fig:learning_filter_bode}
    \vspace{-10pt}
\end{figure}

Fig. \ref{fig:learning_filter_bode} shows the bode plots of designed learning filters for UAV$\#(2)$ learning from UAV$\#(1)$ and UAV$\#(3)$ learning from UAV$\#(2)$. Fig. \ref{fig:te_omega} shows the bode plots for $Te_{1,j}$, and $Te_{2,j}$ (Eq. (\ref{equation:ej_tej})). The bode plots of $\Omega_j$ are also shown in the same figure, where
\begin{equation}
\label{equation:disturbance_estimation_omega}
    \begin{split}
        \hat d'_j &= \Omega_j \{d\}
    \end{split}
\end{equation}
and $\Omega_j$ can be given by
\begin{equation}
        \Omega_j = [1-Q_j+P_j(M_j+C_j)]^{-1}(M_j+Q_jC_j)P_j
\end{equation}

\subsection{Simulation and Experimental Procedure}
\label{subsection:simulation_experiment_methodology}

\begin{figure}[b]
    \centering \includegraphics[width=0.29\textwidth]{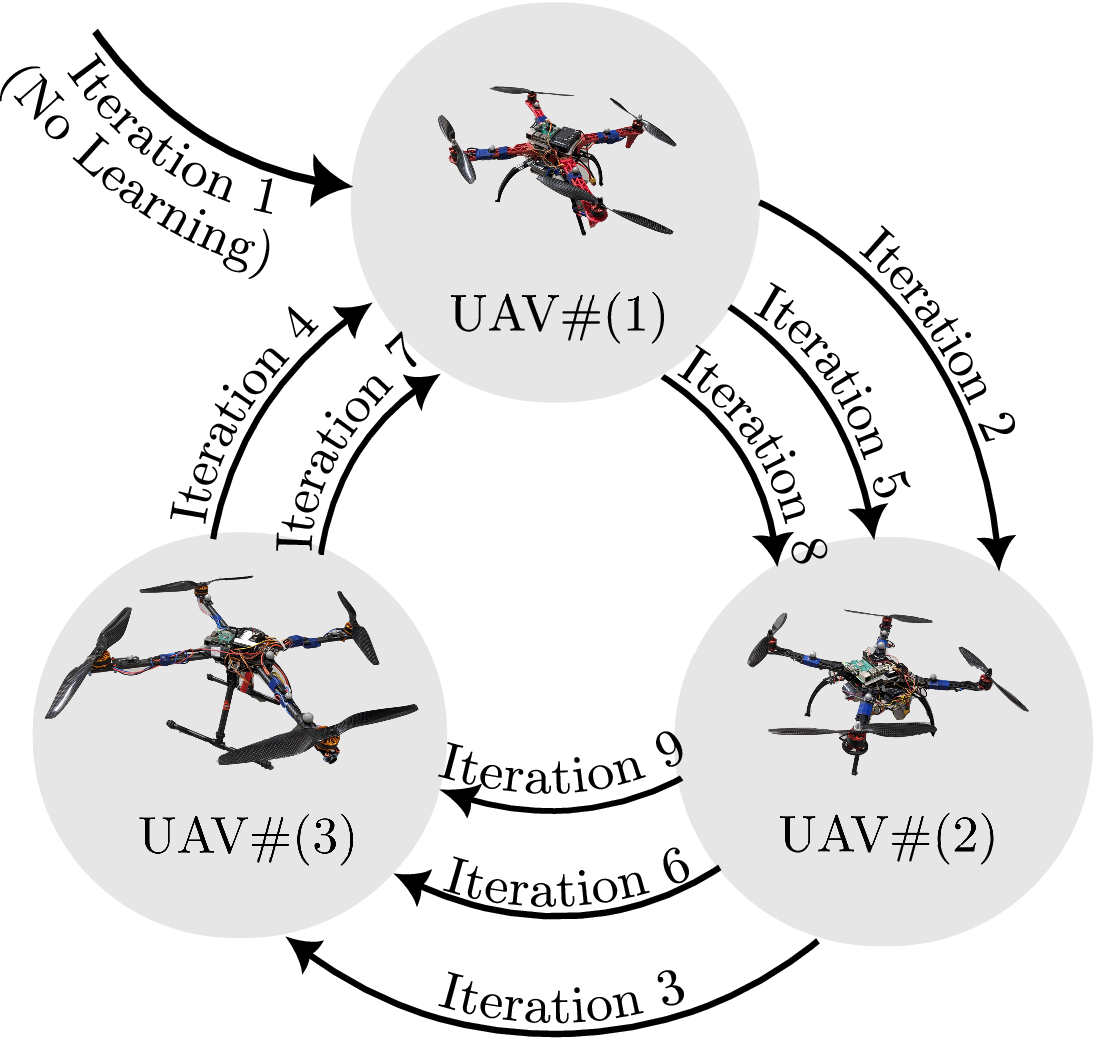}
    \caption{Learning flow for simulations and experiments}
    \label{fig:drones_order}
    \vspace{-5pt}
\end{figure}  

The simulation for all the validations is done in MATLAB using the transfer functions derived from system identification experiments. The Eq. (\ref{equation:yj}) and Eq. (\ref{equation:yj_prime}) are used to simulate the trajectories followed by the UAVs for ``without learning" and ``with learning" cases respectively. Also, as the system parameters for the simulations are accurate, they reflect the ideal case presented in the proof, while the experiments reflect the performance of the learning framework with some deviation from the ideal scenario. For experiments, the UAVs are equipped with a Pixhawk flight controller responsible for controlling the UAV's attitude. Additionally, a Raspberry Pi serves as a companion computer to the Pixhawk. The position of the UAVs is tracked using Vicon motion capture cameras, and the position data is transmitted to the Raspberry Pi. The Raspberry Pi runs a trajectory (outerloop) controller that generates desired attitude and thrust inputs for the Pixhawk's attitude controller. Both the trajectory controller and the attitude controller utilize PID control algorithms. The gains of the PID controllers are not a concern for this study, as long as the system remains stable. The basic PID controller for trajectory control is based on the work presented in \cite{outerloop}. Furthermore, all systems employ the DOB algorithm as part of the trajectory controller. To keep the validation and description neat, the disturbance is added to the x-direction only and thus the learning is just for the x-direction.
To introduce the disturbance into the system, the corresponding acceleration is virtually added to the control signal just before it reaches the plant. Importantly, the disturbance remains unknown to the controller. 

For learning on hardware, the error data and the learning signal from a previous UAV are passed through their respective learning filters and a low-pass filter. This process generates the learning signal for the next UAV. The resulting learning signal is then stored in the next UAV and used by the trajectory controller. Alternatively, the learning signal can also be generated onboard in the next UAV by performing a simple time series conversion. Also, for both the simulations and experiments, the learning is done in a cyclic way (i.e. after system 3, system 1 is tried again with the learning data from system 3) until a convergence is achieved as shown in Fig. \ref{fig:drones_order}. We also perform 2 additional experiments for each UAV for data comparison purposes: (1) without learning and without DOB and (2) without learning but with DOB.

\subsection{Validation Scenarios}

\begin{figure}[!hbpt]
    \centering 
    {\includegraphics[width=0.4\textwidth]        {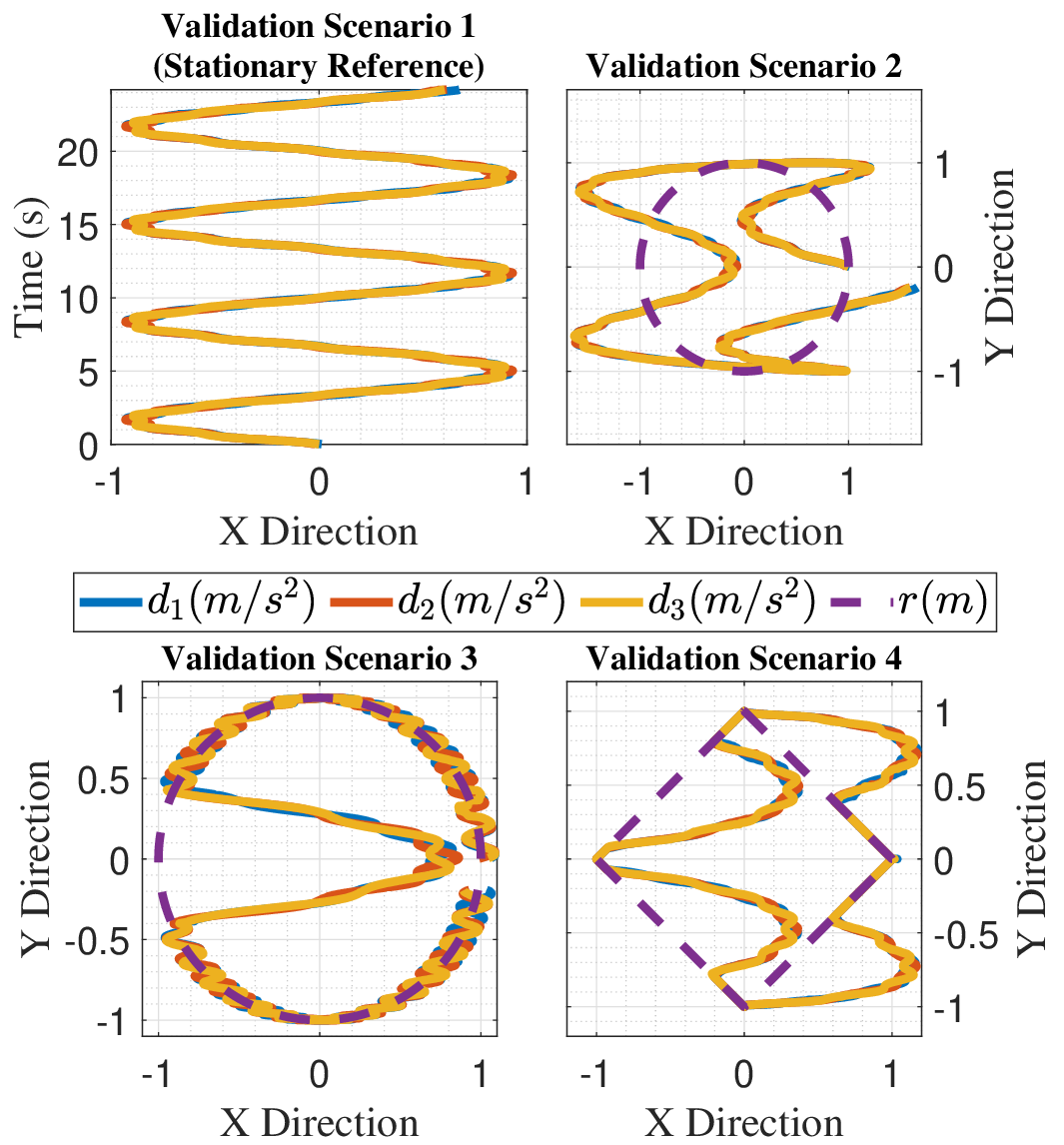}}
    \caption{Reference trajectories (dashed lines) and disturbance profiles (solid lines)}
    \label{fig:scenarios}
    \vspace{-10pt}
\end{figure}

With the simulation and experimental procedure described in Section \ref{subsection:simulation_experiment_methodology}, we assess the effectiveness of the proposed learning methodology with various reference trajectories and disturbance scenarios. We consider 4 distinct reference trajectory-disturbance combinations with increasing complexities as shown in Fig. \ref{fig:scenarios}. The reference trajectory is a stationary point (x: 0 m, y: 0 m, z: 2.5 m) in scenario 1, while it is a circle of radius 1 m with a constant altitude of 2.5 m in scenarios 2 and 3, and a diamond shape of 2 m diagonal length in scenario 4. 

In scenarios 1 and 2, the sinusoidal disturbance of 0.9425 $rad/s$ and amplitude 1 $m/s^2$ is introduced in the x direction. In scenario 3, a half sinusoidal impulse disturbance of the same frequency, but with a larger magnitude of 2 $m/s^2$ is introduced. In scenario 4, a sinusoidal disturbance of 1.4138 $rad/s$ frequency and 1 $m/s^2$ amplitude, but rectified only above 0 is introduced. For each scenario, a sinusoidal noise of larger frequency but smaller amplitude is added to each system differently. This replicates the real-life conditions, in which the disturbance is not exactly the same among the systems.

Consider the frequencies used for the validations with the bode gain of $G_{d,j}$ in Fig. \ref{fig:grd1_grd2_grd3}. For all 3 UAVs, the frequency of 0.9425 $rad/s$ used in scenarios 1 through 3 has a high $G_{d,j}$ gain. This gain is almost maximum at 1.4138 $rad/s$, a disturbance frequency used in scenario 4. These disturbance frequencies for validations are chosen considering this gain as the UAVs' trajectories are most affected by the disturbances at these frequencies, which the learning framework is designed to mitigate. The large magnitude impulse disturbance introduced for scenario 3 replicates a sudden gust of the wind on the UAVs. In scenario 4, the sinusoidal disturbance clipped only in a positive direction replicates the intermittent wind between the buildings. Also, scenario 4 is coupled with sharp corners reference trajectory, which is difficult to track for dynamic systems like UAVs. The UAVs are flown in a different order in scenario 4 (UAV$\#(1)\implies$UAV$\#(3)\implies$UAV$\#(2)$) to test the adaptability of the learning framework. In summary, these scenarios are designed to validate the proposed learning framework rigorously, proving the robustness of the same.

\subsection{Results}
\label{subsection:results}

\begin{figure*}[!htpb]
    \centering 
    \includegraphics[width=1.0\textwidth]{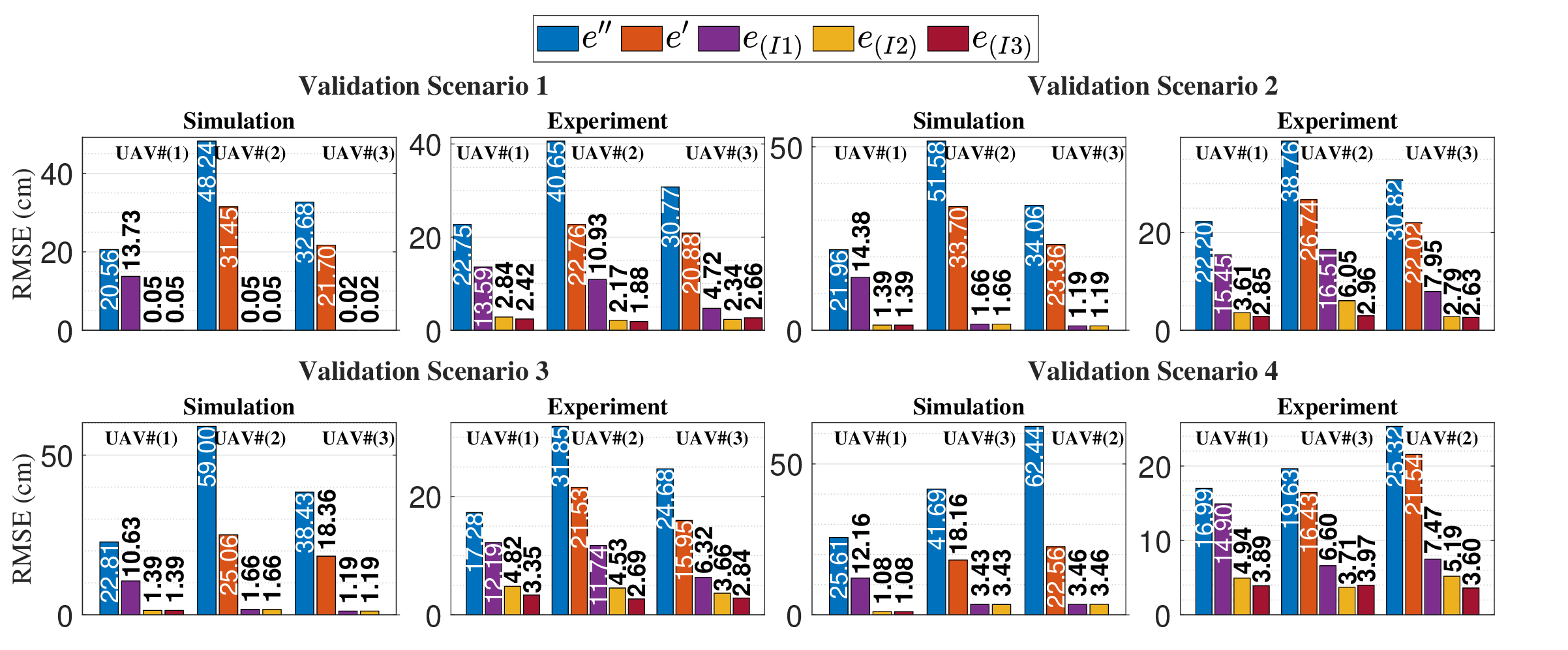}
    \caption{RMSE of the trajectory tracking for all the scenarios in both the simulations and experiments. $e''$ indicates the ``no DOB" case, $e'$ indicates the ``only DOB" case and $e_{(Ik)}$ indicates the $k$\textsuperscript{th} iteration for each system.}
    \label{fig:errors_RMSE_8}
    \vspace{-10pt}
\end{figure*}  
   
\begin{figure*}[!hbpt]
    \centering 
    {\includegraphics[width=1.0\textwidth]{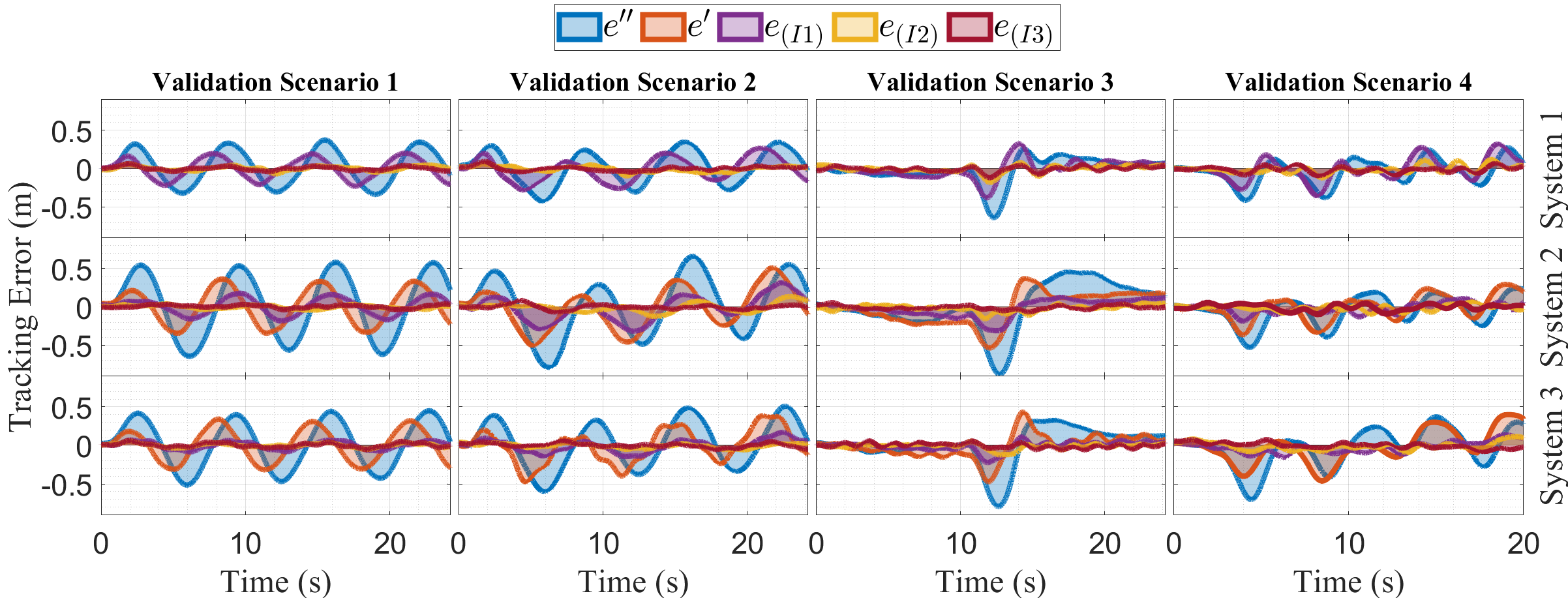}}
    \caption{Trajectory tracking errors in experiments for all 4 Scenarios. $e''$ indicates the ``no DOB" case, $e'$ indicates the ``only DOB" case and $e_{(Ik)}$ indicates the $k$\textsuperscript{th} learning  iteration for each system.}
    \label{fig:tracking_errors_4}
    \vspace{-10pt}
\end{figure*}

\begin{figure*}[!htpb]
    \centering 
    \includegraphics[width=1.0\textwidth]{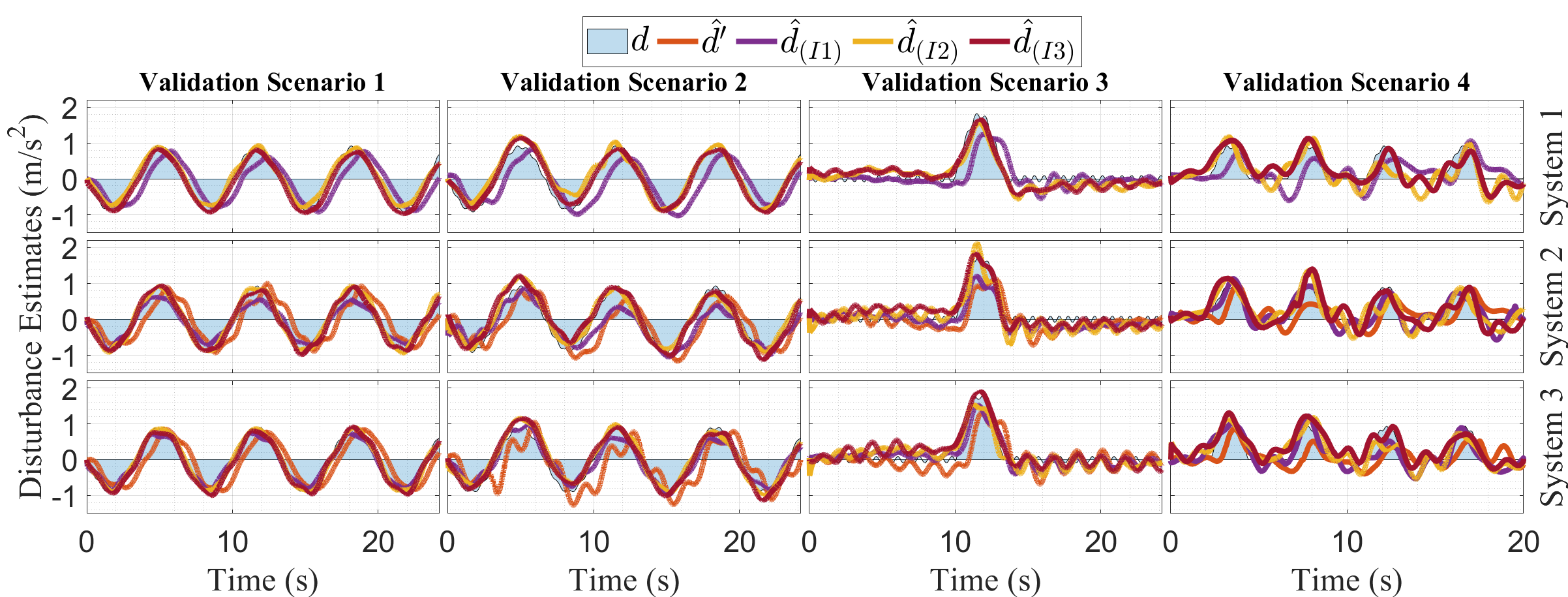}
    \caption{Disturbance estimates for experiments in all the scenarios. $d'$ indicates the disturbance estimates in the ``only DOB" case and $d_{(Ik)}$ indicates the disturbance estimates with learning in $k$\textsuperscript{th} iteration for each system.}
    \label{fig:disturbance_estimations}
    \vspace{-10pt}
\end{figure*}  

In this section, we present detailed results from the validation simulations and experiments. The video from the experiments is available at \href{https://zh.engr.tamu.edu/wp-content/uploads/sites/310/2024/02/ILCDOB_v3f.mp4}{Link}. Fig. \ref{fig:errors_RMSE_8} shows the Root Mean Square Error (RMSE) of the trajectory tracking for each scenario in both simulations and experiments. The RMSE of the trajectory tracking reflects the summary of each experiment so that we can quantitatively compare them. For each scenario and each system, we compare the errors in ``No DOB" cases (blue bars), ``Only DOB" cases without learning (orange bars), and the errors in $1^{st}$, $2^{nd}$, and $3^{rd}$ learning iterations of each system. Please note that we are comparing the results between various experiments within the systems and not with other systems as differences in system dynamics affect the performance. For all the validation scenarios, the trajectory tracking error without learning is maximum, which reduces marginally with DOB. However, the errors with learning reduce significantly for each system. For simulations, the errors with learning converge in the first learning iteration itself as the system models are accurate. For experiments, we can observe that in the $2^{nd}$ and the $3^{rd}$ learning iterations, the errors are very close to 0 for each system. 

Expanding further into the summary of Fig. \ref{fig:errors_RMSE_8}, Fig. \ref{fig:tracking_errors_4} shows the trajectory tracking errors in the experiments for all the validation scenarios against time. Each row of plots refers to a particular system whereas each column of plots refers to a validation scenario. Here also, it is evident that with DOB but without learning, the errors are marginally reduced compared to ``without DOB" cases. With learning, the errors reduce significantly compared to the ``only DOB" cases, which become almost negligible in the $2^{nd}$ and the $3^{rd}$ iterations.

The learning framework also improves the disturbance estimates as shown in Fig. \ref{fig:disturbance_estimations}. The shaded shapes on each plot reflect the actual disturbances while the lines show the estimated disturbances. We can observe that in the ``only DOB" cases, the disturbance estimates in scenarios 1 and 2 are almost perfect, but are delayed. This is expected as the DOB estimates the performance using the output of the system, which is delayed compared to the disturbances. In scenario 3, the estimates are delayed as well as scaled down in the ``only DOB" cases. In scenario 4, the estimates are highly inaccurate in the ``only DOB" cases, possibly because the disturbance was clipped in a positive direction. However, with the first iteration of learning, the estimates are improved significantly, which explains why the trajectory tracking performance is better with learning. In $2^{nd}$ and $3^{rd}$ learning iterations of each system, the disturbances are nearly fully recovered, which
validates the effectiveness of the learning framework.

\section{Conclusions}
\label{section:conclusion}
This study presents a framework for designing ILC with DOB in systems with mismatched dynamics. Both numerical and experimental validations demonstrate the effectiveness of the framework and the learning filters in reducing the tracking error and improving the disturbance estimate under various disturbance scenarios, enhancing the system's robustness to external disturbances. Future research directions include exploring learning with highly aggressive reference trajectories. Additionally, the current approach utilizes errors and learning signals from only the previous system for learning. To further improve the learning framework, future work can explicitly incorporate data from all previous systems into the learning process, rather than relying solely on the implicit information contained in the learning signal of the previous system.

\end{document}